# Text line Segmentation in Compressed Representation of Handwritten Document using Tunneling Algorithm

Amarnath R.*[1], Nagabhushan P. [1]

*Abstract:* In this research work, we perform text line segmentation directly in compressed representation of an unconstraint handwritten document image using tunneling algorithm. In this relation, we make use of text line terminal point which is the current state-of-the-art that enables text line segmentation. The terminal points spotted along both margins (left and right) of a document image for every text line are considered as source and target respectively. The effort in spotting the terminal positions is performed directly in the compressed domain. The tunneling algorithm uses a single agent (or robot) to identify the coordinate positions in the compressed representation to perform text-line segmentation of the document. The agent starts at a source point and progressively tunnels a path routing in between two adjacent text lines and reaches the probable target. The agent's navigation path from source to the target bypassing obstacles, if any, results in segregating the two adjacent text lines. However, the target point would be known only when the agent reaches destination; this is applicable for all source points and henceforth we could analyze the correspondence between source and target nodes. In compressed representation of a document image, the continuous pixel values in a spatial domain are available in the form of batches known as white-runs (background) and black-runs (foreground). These batches are considered as features of a document image represented in a Grid. Performing text-line segmentation using these features makes the system inexpensive when compared to spatial domain processing. Artificial Intelligence in Expert systems, dynamic programming and greedy strategies are employed for every search space while tunneling. An exhaustive experimentation is carried out on various benchmark datasets including ICDAR13 and the performances are reported.

## 1. Introduction

Technological advances of storage and transfer have made it possible to maintain many documents in the digital format. It is also necessary to preserve these documents in digital image format only, particularly in case of handwritten documents for verification and authentication purposes [1, 2, 3, 4, 5]. Maintaining these document images in the digital form would require huge storage space and network bandwidth. Therefore, an efficient compressed representation would be an effective solution to the storage and transfer issues [6, 7, 8] particularly arising from document image. Generally, the document image compression format follows the guidelines of CCITT Standards [9, 10] which is a part of the ITU (International Telegraph Union). These standards were specifically targeted towards document images stored in digital libraries. On the other hand, if document images, audios and videos are to be archived and communicated in the compressed form itself, then it would be considered as a third advantage in addition to storage and transmission.

The digital libraries with document images in their compressed formats could imply a solution to a big data problem arising from the document images, particularly about storage and transmission [11]. The compressed image file formats such as TIFF, JPEG, and PNG, strictly follow CCITT standards [7, 8, 9, 10, 11, 12].

For any digital document analysis (DDA), the image in its compressed format must undergo the decompression stage. However, performing operations on decompressed documents would unwarrantedly suppress the advantages of both the time and buffer space [6, 13, 14, 15]. If DDA could be achieved in the compressed version of the document image without decompression, then the document image compression could be viewed as an effective solution to the immense data problems arising from document images [13].

The idea of operating and analysing directly in the compressed version of the document image without decompression is known as Compressed Domain Processing (CDP) [6, 13, 14, 15). A recent literature [12] on CDP shows the strategies to perform document image analysis in its compressed representation. However, the model is subjected to the printed documents only. The challenging job is to perform DDA in compressed representation of handwritten document image. Performing DDA on uncompressed handwritten document could be a difficult task because of oscillatory variations, inclined orientation and frequent touching of text lines while scribing on un-ruled paper [13]. An initial attempt [13] to spot the separator points at every line terminals in compressed unconstrained handwritten document images using run-length features (Run Length Encoding or RLE) is the state-of-the-art technology. This motivates to carryout text-line segmentation in compressed document image. In this research, we have attempted to segment the text lines in the compressed representation with the help of these spotted terminal points. In this context, a research [16] shows a method of extracting run-length features from the compressed image format supported by CCITT group 3 and CCITT group 4 standards. These protocols use run-length encoding, which is widely accepted for binary document image compression. Our paper aims to segment the text lines of the document image using the run-length data which is represented in a grid. The detailed explanation about the RLE format of the document image is provided in Section 3.

[1] *Department of Studies in Computer Science, University of Mysore, Karnataka, India*
*\* Corresponding Author Email: amarnathresearch@gmail.com*

The current state-of-the-art uses a single column (first column) of the grid to find the separator points at every line terminal along left margin of the document image. But the last column of RLE does not infer the depth of right margin of the document for spotting separator points. To work on the right margin of the document image, the final non-zero entry of every row in RLE data was observed for creating a virtual column and thus separator points are spotted. Now, these identified separator points at both the margins are considered as source (separator points along left margin) and target (separator points along right margin) nodes. In this paper, we make use of single agent (or robot) for text-line segmentation. The agent's job is to tunnel or trace the path by navigating between the two-adjacent text-lines starting from a source point at one end of a document and reaching the destination point present at the other end of the document. This process is applied to all the source nodes resulting in segmentation of all text-lines. Some interesting observations inculcated includes analysing the correspondence between the source nodes and terminal nodes. We also addressed some of the issues related to wrong segmentation that is when two or more paths converge to one destination traced by the agent. Corrective measurements are taken to tackle this issue by understanding the correlation between two adjacent paths.

Our approach identifies the text line positions operating directly in the RLE representation of the document images which results in text-line segmentation. The rest of the paper is organized as follows. A survey of related research is presented in section 2. In section 3, we have explained the RLE representation of a document image. Further, we provide the terminologies and corollaries used in this paper. Section 4 describes the algorithmic modelling along with comparative time complexity with respect to both RLE and uncompressed document versions. Further, experimental results conducted on benchmark handwritten datasets such as ICDAR 13 and other databases [17] are described in Section 5. Section 6 summarizes the research work with future avenues.

## 2. Related Works

In the recent past, we could trace few related works on CDP, but restricted to printed document images. A literature [15] on CDP provides a detailed study on document image analysis techniques from the perspectives of image processing, image compression and compressed domain processing. This enabled various operations carried out in the field of skew detection / correction, document matching, document archival.

Recent article [18], demonstrates straight line detection in RLE representation of the handwritten document image. Incremental learning-based text-line segmentation in compressed handwritten document images are reported in literature [19]. Further, there was a technical article [13] that discusses about performing direct operations on the compressed representation of handwritten document. The effort is to spot the separator points at every text lines in both margins of the document image enabling text line segmentation. For better understanding, the identified line terminal points applied over an uncompressed document image is shown in Figure 1.

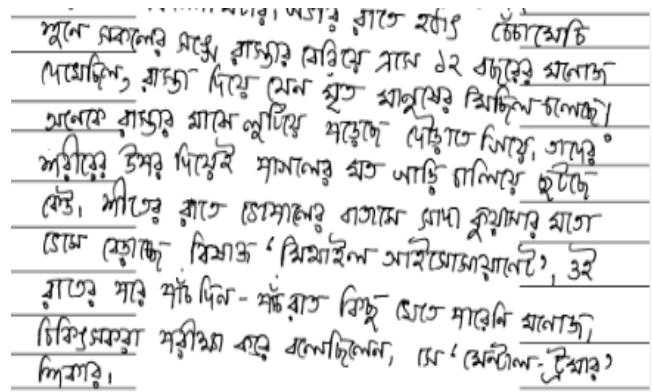

*Fig. 1. Line Separators at both borders of a document in uncompressed documents* (Reference: A portion of ICDAR13 test image - 320.tif)

One of the significant models [20] uses seam carving approach to segment the text-line that works on historical documents of uncompressed images. Another recent study [21] discusses various path finding algorithms which are a class of heuristic algorithms based on Artificial Intelligent. These techniques are domain specific that works on uncompressed document images. However, the ideas of path finding approaches are inculcated in our proposed model, that is to operate on compressed version for decision-making in every search space. Further, an effort was made to carry out the text-line segmentation directly in compressed handwritten document images to avoid decompression [13] which is the main objective of this paper.

## 3. Compressed Image Representation and Corollaries

The Modified Huffman (MH) [7, 8] is the most common encoding technique following CCITT compression standards which is supported by all the facsimiles. The improved compression versions are Modified Read (MR) [7, 8, 9, 10] and Modified Modified Read (MMR) [7, 8, 9, 10]. A comparative study on encoding / decoding techniques of these compression standards are tabulated (Table 1). MH uses RLE for an efficient coding, whereas MR and MMR exploit the correlation between successive lines of a document image.

**Table 1.** Compression standards

| Characteristics | MH | MR | MMR |
|---|---|---|---|
| CCITT Standard | T.4 | T.4 | T.6 |
| Dimension | 1-D | 2-D | 2-D (Extended) |
| Algorithm | Huffman and RLE | Similarity between two successive lines | Extended MR |

Figures 2, 3 and 4 show a sample binary image, its corresponding hexadecimal image viewer and the RLE representation of the image respectively. The RLE representation of an image is defined as Grid (G) in this paper.

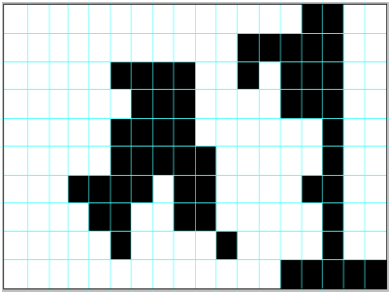

**Fig. 2.** A sample image of size 18X10 (WidthXHeight)

```
Hexadecimal
FF D8 FF E0 00 10 4A 46 49 46 00 01 02 01 01 2C
01 2C 00 00 FF E1 04 04 45 78 69 66 00 00 4D 4D
00 2A 00 00 00 08 00 07 01 12 00 03 00 00 00 01
00 01 00 00 01 1A 00 05 00 00 00 01 00 00 00 62
01 1B 00 05 00 00 00 01 00 00 00 6A 01 28 00 03
00 00 00 01 00 02 00 00 01 31 00 02 00 00 00 1C
00 00 00 72 01 32 00 02 00 00 00 14 00 00 00 8E
87 69 00 04 00 00 00 01 00 00 00 A4 00 00 00 D0
00 2D C6 C0 00 00 27 10 00 2D C6 C0 00 00 27 10
41 64 6F 62 65 20 50 68 6F 74 6F 73 68 6F 70 20
43 53 33 20 57 69 6E 64 6F 77 73 00 32 30 31 37
3A 30 39 3A 31 32 20 31 33 3A 3A 34 33 3A 35 34 00
00 00 00 03 A0 01 00 03 00 00 00 01 00 01 00 00
A0 02 00 04 00 00 00 01 00 00 00 12 A0 03 00 04
00 00 00 01 00 00 00 0A 00 00 00 00 00 00 00 06
01 03 00 03 00 00 00 01 00 06 00 00 01 1A 00 05
00 00 00 01 00 00 01 1E 01 1B 00 05 00 00 00 01
00 00 01 26 01 28 00 03 00 00 00 01 00 02 00 00
02 01 00 04 00 00 00 01 00 00 01 2E 02 02 00 04
00 00 00 01 00 00 02 CE 00 00 00 00 00 00 00 48
00 00 00 01 00 00 00 48 00 00 00 01 FF D8 FF E0
00 10 4A 46 49 46 00 01 02 00 00 48 00 48 00 00
```

**Fig. 3.** Hexadecimal visualization of the image from Fig. 2

| 14 | 2 | 2 | 0 | 0 | 0 | 0 |
|----|---|---|---|---|---|---|
| 11 | 5 | 2 | 0 | 0 | 0 | 0 |
| 5  | 4 | 2 | 1 | 1 | 3 | 2 |
| 6  | 3 | 4 | 3 | 2 | 0 | 0 |
| 5  | 4 | 6 | 1 | 2 | 0 | 0 |
| 5  | 5 | 5 | 1 | 2 | 0 | 0 |
| 3  | 4 | 1 | 2 | 4 | 2 | 2 |
| 4  | 2 | 2 | 2 | 5 | 1 | 2 |
| 5  | 1 | 4 | 1 | 4 | 1 | 2 |
| 13 | 5 | 0 | 0 | 0 | 0 | 0 |

**Fig. 4.** RLE data inferred as Grid of the image from Fig 2

The odd and even columns of the RLE data represent the count of white-runs and the black-runs of an uncompressed document image. The number of continuous pixel value, say '0' (background), is called as white-runs [22, 23, 24, 25, 26]. Similarly, the number of continuous pixel value, say '1' (foreground) is called as black-runs [22, 23, 24, 25, 26]. If a row in an image starts directly with foreground, then the value in the first column of the corresponding row in RLE has an entry zero ('0'). This infers that odd columns and even columns in the grid contain count of white-runs and black-runs respectively. The white-runs represent background whereas the black-runs represent foreground as text-content.

Some of the corollaries used in this article are listed here:
a. Each separator points spotted in the first column of the Grid ($G$) is considered as Source Node ($S$).
b. Each separator points spotted in the virtual column is considered as Target Node ($T$). Final non-zero entry in every row of G makes this virtual column.
c. A path ($P$) is defined as segmentation of adjacent text lines; an agent (or robot) navigates along blank space (white-runs) available between two adjacent text-lines from $S$ to $T$ with reference to $G$.
d. Edges ($E$) are defined as sequences of white-runs that exist or tunneled between two adjacent text-lines. These sequence of $E$ formulates $P$.
e. An obstacle ($O$) appears because of touching characters between two adjacent text-lines, which is a hurdle while tracing $P$ from $S$ to $T$. This may also occur due to ascenders and descenders of characters, which appears larger in length when compared to a given threshold ($t$).
f. $t$ is the length covered by an agent tracing both the direction vertically up ($j-t$) and down ($j+t$) from the current position $G(j,i)$, where arguments $j$ and $i$ refer to the position in $G$ along $y$ (columns) and $x$ (rows) axis respectively; this search space is induced whenever there is an $O$. $t$ is chosen empirically based on experiments conducted on the datasets.
g. For every $S$, there is a $T$. The relationship between $S$ and $T$ is one-to-one function defined as $f(S) = f(T)$ and strictly follows the ascending order. The correspondence of $S$ and $T$ is not defined in the literature [13]. However, in this paper, we could find the probable correspondence between $S$ and $T$ with help of an agent (or robot) and algorithmic strategies.
h. When relationship between $S$ and $T$ is onto function or crossover, then it is observed as wrong segmentation. Even though $S$ and $T$ hold a relation one-to-one and there exist crossovers, then it is considered as wrong segmentation.
i. The distance ($d$) calculated between $S$ and $T$ denotes the shortest $P$ covering longest distance (weights), which is defined by a function, $d(S,T)$. $d$ depends on the number of edges ($E$) used in $P$ by an agent to cover $P$ between $S$ and $T$.
j. The terminal points $S$ and $T$ does not necessarily possess larger white-runs because, they are heuristically chosen as mid-points of the bands corresponding to the two-adjacent text-lines along $y$ axis (columns) of $G$. Therefore, $S$ may not be an actual start point, in the given situation, whereas the weight of the $S'$ is longer than $S$. So, the actual start point would be $S'$ and it would be within the search space of $t$ from $S$.
k. The distance $d(h, h')$ between two intermediate hubs such as $h$ and $h'$ must be equal or greater than a given threshold ($t'$). $t'$ is calculated by finding a maximum number of bins observed from the histogram of $G$ with respect to odd columns (white-runs) only.

Some of the assumptions concerning the state space search (tunnel or traversal or $P$) are given below:
 a. $|S|$ and $|T|$ are finite.
 b. There must be at-least one $P$ between $S$ and $T$
 c. One agent (or robot) is employed at a time for tracing $P$.
 d. There is no self-loop in $SS$ (a cycle of length one).
 e. Tunnel (Move to next state) if $O$ exist on $P$.

Figure 5 shows the terminal points of $S$ and $T$ in an uncompressed version for better understanding. It depicts 10

source nodes and 10 target nodes. The one-to-one function for the figure 5 is defined below:

$Domain, S = \{S1, S2, ..., S10\}$

$CoDomain, T = \{T1, T2, ..., T10\}$

$f: S \rightarrow T$

$f(S1) = T2, f(S2) = T3, ..., f(S9) = T10$ and $f(S10) = Empty$

$S1 < S2 < S3 < \cdots < S10$

$Similarly, T1 < T2 < T3 < \cdots < T10$

$If\ f(S_v) = T_u\ then\ strictly\ f(S_{v-1}) = T_{u-1}\ and\ f(S_{v+1}) = T_{u+1} \forall v\ and\ u$

Based on the observation, the source $S10$ does not have a corresponding target $T$. Therefore, an agent starting from $S$ may reach $T$ or $\hat{T}$, where $\hat{T}$ is presumed as a new $T$ (correspondence) or $T'$. This infers that the agent may reach closer or nearer to $T$ but not exactly the target $T$, which is illustrated in the following section.

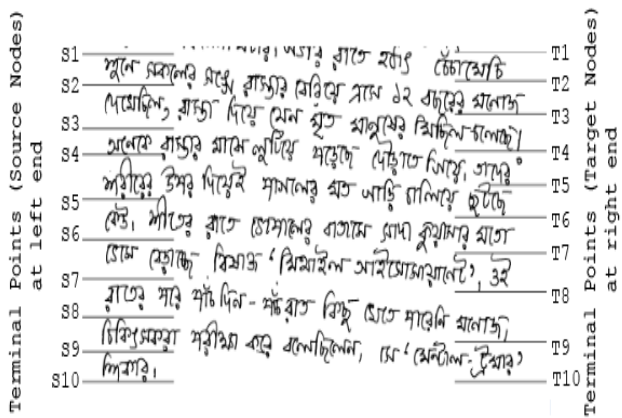

**Fig. 5.** Line Separators displaying S and T at both ends of a document in uncompressed documents. (Reference: A portion of ICDAR13 test image - 320.tif)

## 4. Text-line Segmentation

Grid ($G$) of a document image starts with a column of white-run(s). The starting column of $G$ is referred [13] as a white space that exists at the beginning of a text document. A larger depth (white-runs) in this column indicates the text-line separation gap (bands) along left margin of the document image between the corresponding adjacent text-lines. Mid points are chosen as terminal points (Source nodes) from the constructed bands (corollary $j$) as shown in the figure 5. In this context, these points do not necessarily possess larger white-runs. Therefore, the agent needs to choose the starting point $S$ among its vertical neighbors having largest white-runs among its vertical neighbors within the range $t$ from its current position. Now, assuming that the agent's start position S is fixed based on the corollary $f$ and this state is at an initial search state $SS(0)$. Next, the agent reaches at a station or hub ($h$) starting from $S$ resulting in a new search state $SS(1)$. An edge $E$ connecting between these two stations $S$ and $h$ is the shortest path travelled by the agent to cover maximum distance and obviously heading towards the probable target. The distance or weight, $d(S, h)$ from $S$ to an intermediate hub $h$ is computed directly from the position, say $G(j, 1)$, then it is inferred in the first column of $G$. The search process continues until the agent reaches the corresponding $T$ or new target $T'$ (probable destination).

The basic idea is that for every $SS$, the agent needs to choose the next hub which is nearer to $T$ or $T'$ by selecting the white-runs from the current position in $G$ and thus ensures that a direct edge $E$ exists between the two stations (hubs). In other words, the agent selects the largest white-runs from $G$ to adhoc, to navigate to the next hub closer to $T$ or $T'$. At every $SS$, greedy strategy is employed to identify the next successive hub h' by choosing largest white-runs (or largest weight) from the current $h$ and henceforth the agent reaches $T$ or $T'$ using a minimum number of $E$ with respect to the hubs visited. The agent hopping hubs starting from $S$ to $T$ or $T'$ results in creating a path $P$. $P$ is based on the edges connected between the intermediate hubs starting from $S$ to $T$ or $T'$ and this indicates a progressive segmentation of the two-adjacent text-lines in a document image, starting from the left margin and precedes towards the right direction. If the total search space for a given $G$ is one, then this indicates that the agent has visited only two hubs and those are $S$ and $T$ respectively; this articulates direct segmentation without much variations exist because of skew, curvy and touching lines in the handwritten document image. This situation sometimes would be pronounced in-between two paragraphs in the document image holding a large horizontal space between them and may occur in the top and bottom of the documents which is entitled to have an adequate margin spaces. However, the whole operation is carried out with reference to $G$.

Consider that the agent reaches an intermediate hub which is at $SS(1)$, then we perform local search by employing corollary $f$ to reach to the next hub ($h'$). This situation is related to Hill-Climbing [27] technique in Artificial Intelligence. If the hubs h and h' visited by the agent are located in the positions say $G(y', x')$ and $G(y'', x'')$ respectively, then the search space (local maxima) is written as.

$$h^* = SS(>0) = \sum_{i=1}^{n} G(j, i), when\ G(j, i) > 0\ \&break\ if\ Cumulative\ Sum > d(S', h^{*-1}),$$

$where (y - t) \geq j \leq (y + t)$

$where\ n = number\ of\ columns\ in\ G\ and\ * \ refers\ to\ next\ hub$

The selection of each successive hub is based on the distance d computed by summing up the entries from successive columns in $G$ along $x$ axis starting from the first column with respect to the y coordinate position. The distance computation is carried out for every $y$ coordinate position which varies within the range of $t$ from either side of position, $j$, as given in the above equation. The job of the agent is to select the largest weight (distance) among these computed cumulative values which crosses the current hub (intermediate hub). However, measuring the distances for every $SS$ ranging between $j - t$ to $j + t$ (both inclusive) along $y$ axis is computationally expensive. In computing, memoization [28] or memorize [28] is an optimization technique used primarily to speed up computer programs by storing the results of expensive function calls and returning the cached result when the same inputs occur again. To avoid this re-computation for every $SS$, we use memoization or memoize technique in dynamic programming strategy to remember the last computed $d$ of previous $SS$ which in-turn reduces the computational complexity. The gap or distance $d(h, h')$ between two successive hubs must be within a minimum distance of a given threshold $t'$. $t'$ is chosen heuristically based on the experiments conducted on the benchmark datasets.

A common issue encountered as mentioned in the corollary $e$ is when two adjacent text-lines are touching one another. In this case, we just bypass or crossover the lines [29]. Algorithmically, we choose next successive column (black-runs) with respect to the current position. In other case, the same strategy is extended when the ascenders and descenders appear to be greater than the given threshold value $t$, where the agent would not be able to go beyond the search limit, say $t$. One of the important issues mentioned in corollary $k$ is when a distance $d(h, h')$ between hubs is greater than $t'$. $t'$ is given below.

*t'= maximum(histogram(values in odd colums of G))*
*where values in odd columns of G represents white runs*

In this case, we calculate $t'$ by considering the largest bin from the histogram of $G$ with respect to the counts of white runs (odd columns). If the distance is found to be larger than $t'$, then we apply both the concepts namely 'don't care condition' [29] and 'minimal cut-sets' [30] from graph theoretical domain. Therefore, with a minimal number of crossovers, the agent bypasses and reaches the next successive hubs. Finally, the agent will update its knowledge base about each visited hub with reference to $G$ covering the distance from $S$ to $T$ or $T'$. Below we present the proposed algorithm for text-line segmentation using $G$.

*Algorithm: Finding text-line segmentation in $G$ using tunneling approach*

**Input:** Compressed representation $G$ of a document image, Terminal points – Source Nodes ($S_n$) at ever line terminals. Vertical neighboring search threshold $t$.

**Output:** Text-line segmentation in $G=TS$

1. For-each Source Node ($S$) from $S_n$, with the position $G(y',x')$, where, $x'=1$;

*CurrentSum=0;*
*Maximum=0;*
*GridY=y'*

While $j=G(y'-t,x')$ to $G(y'+t,x')$ the agent vertical search range is between $y'-t$ to $y'+t$ (both inclusive), where $j>=1$ and $j<=high$ of $G$.

a. Calculate: Cumulative Sum (Initialize $Csum=0$) for each $j$ or Use Dynamic Programming. Use Memoize to check data in Knowledge base (KB)

$$\text{if } KB(j) \text{ is available}$$
$$S(j) = KB(j) + \sum_{i=p\,from\,KB}^{n} G(j,i),$$
$$\text{when } G(j,i) \neq 0 \text{ and}$$
$$Csum(S(j)) > CurrentSum$$

$$\text{Else}$$
$$S(j) = \begin{cases} G(j,1), \text{when } x' = 1 \\ \sum_{i=1}^{n} G(j,i), \text{when } G(j,i) \neq 0 \text{ and} \\ Csum(S(j)) > CurrentSum \end{cases}$$

$$Memory(j) = S(j)$$
$$End\ if$$

a. Find Maximum and update $y$ coordinate position:

$$if\ S(j) > Maximum$$
$$Maximum = S(j)$$
$$GridY = j$$
$$GridX = i$$
$$End\ if$$

2. Update $x$ and $y$ coordinate position of $G$:
$$New\ postion, y' = GridY$$

$$Travel\ distance\ d = Maximum$$

The positions chosen in the grid is $G(GridY, GridX)$ update in $TS$

Goto Step 2 until $Maximum = Total\ Distance\ d(S,T)$

3. **Stop:** When the agent covers the distance $d$.

### 4.1. Time Complexity for Text-line Segmentation

The following provides the time complexity estimated for the proposed algorithm to identify the text-line segmentation positions in compressed representation of a document image. The worst-case scenario is calculated as $O(m' \times n')$ where $m'$ and $n'$ represent the number of rows and number of columns of the grid respectively.

Usage of memoize technique [28] in dynamic programming strategy has reduced to the complexity $O(m' \times n')$ from $f: X \to X$, that is $f: O(m' \times n') \to O(m' \times n')$ which is a non-deterministic polynomial.

For comparative study, the time complexity is also estimated for a document image in its uncompressed format. For this, the worst-case scenario is $O(m \times n)$ where $m$ and $n$ represent the height and width of an uncompressed image respectively. It is noted that $m'$ is equal to $m$ whereas $n'$ is lesser than $n$.

Time complexity of spotting the line terminal points mentioned in the literature [13] is estimated as $O(2 \times m' \times 1)$ in the worst-case scenario. Therefore, the complexity for the proposed algorithm is computed by cascading the two algorithms which result in $O(2 \times m' \times 1) + O(m' \times n')$. Finally, the proposed method takes $O(m' \times n')$.

The overall computation is tabulated (Table 2).

**Table 2.** Time Complexity

| Complexity Worst Case | Compressed Domain | Uncompressed Domain |
|---|---|---|
| Spotting Line Separators | $O(m')$ | $O(mXn)$ |
| Text-line Segmentation without dynamic programming | $O(m'Xn')!$ | $O(mXn)!$ |
| Text-line Segmentation (Proposed method) | $O(m'Xn')$ | $O(mXn)$ |
| Overall Efficacy | $O(m'Xn')$ | $O(mXn)$ |

The efficacy of the proposed algorithm for both compressed and uncompressed versions is shown in Figure 6. The compressed version requires less time, when compared to that of uncompressed images for text line segmentation.

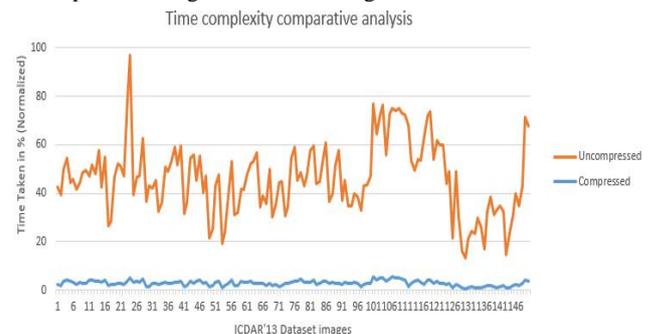

**Fig. 6.** Comparative analysis of both uncompressed and compressed version of document images (Ref: ICDAR'13 test datasets)

**4.2. Illustration**

In this section, we illustrate the working principle of Tunneling algorithm using an example. The coordinate positions are tracked by the agent (or robot) for text line segmentation with respect to $G$ is shown in Table 3.

**Table 3.** RLE coordinate positions (Reference: A portion of ICDAR13 test image – 320.tif – text-lines 21 and 22)

| Hubs (Stations) | Y-Coordinate | X-Coordinate | Distance From Source Node |
|---|---|---|---|
| Source | 1977 | 1 | 767 |
| Hub 1 | 1984 | 1 | 1058 |
| Hub 2 | 2000 | 9 | 1295 |
| Hub 3 | 2020 | 59 | 1325 |
| Hub 4 | 2014 | 53 | 1626 |
| Hub 5 | 2029 | 67 | 1901 |
| Hub 6 | 2049 | 77 | 2085 |

For better understanding, an experimental result is shown in the uncompressed version as well. Figure 7 shows segmentation of two text-lines of a document image. Figure 8 illustrates the text-line segmentation depicting transition hubs and a path tunneled by an agent.

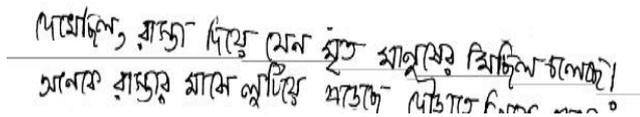

*Fig. 7. Segmentation of two text lines depicted in uncompressed documents* (Reference: A portion of ICDAR13 test image – 320.tif – text-lines 21 and 22)

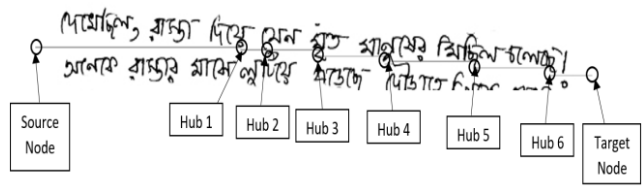

*Fig. 8. Segmentation of two text lines depicted in uncompressed documents* (Reference: A portion of ICDAR13 test image – 320.tif – text-lines 21 and 22)

From the example, the agent starts from the Source Node ($S$) and reaches the Target Node ($T$) visiting six intermediate hubs such as Hub 1, Hub 2, Hub 3, Hub 4, Hub 5 and Hub 6. Here, the source node is identified in a position, say $G(1977,1)$. The agent predicts that the Hub 1 in the position $G(1984,1)$ could be an effective source node than the initial source [13] as articulated in corollary $j$. This is because the weight (distance) of Hub 1 is larger when compared to the initial source as illustrated in table 3. The total distance covered by the agent is 2085 units. The threshold $t$ is chosen as 20 units and the total computation to cover the distance along $y$ coordinates in both the direction by the agent ranges from $1957\ (Min(Y-Coordinate) - t)$ to $2069\ (Max(Y-Coordinate) + t)$, i.e., $Source - t$ to $Hub6 + t$ respectively. We employ memoization technique to reduce the complexity. Therefore, we make use of a knowledge base ($KB$) which would be empty initially. Once we compute the distances for both Source and Hub 1, we feed all the weights (distances) ranging from $G(Source - t, 1)$ to $G(Source + t, 1)$, i.e., $G(1957,1)$ to $G(1998,1)$ respectively into the KB. The size of the KB would be a column size of $G$. The $SS$ triggers the agent to choose next hub as Hub 2 as a successor of the Hub 1 which covers the distance of 1295 units and perhaps the distance is greater than the distance of current hub, that is the Hub 1 possess 1058 units. Next, agent archives the information from the KB to avoid the total computation.

**4.3. Choosing of Source Node**

In this section, we introduce a mechanism to find the start node. Figures 9 (a) and 9(b) show a sample image and its compressed representation in $G$ respectively. Here, the source node $G(6,1)$ is not an optimal start point as mentioned in corollary $j$. If we consider the threshold $(t = 6)$ for illustration purpose, then the search scope will be of range covering from $G(6 - 6, 1)$ to $G(6 + 6, 1)$. Though the position $G(0,1)$ is extended beyond the grid boundary as described in the algorithm in step 2, we re-define the search range starting from $G(1,1)$. Because of $SS$, the two positions such as $G(11,1)$ and $G(12,1)$ are identified as the largest among the search range. In this case, we take the source node $S'$ that is in the position $G(11,1)$ than latter, because $G(11,1)$ is nearer or closer to the initial source node which is $G(6,1)$. Finally, we choose $G(6,1)$ as a start node, and thus the distance covered by the agent would be 7 units and holding a largest weight covering maximum distance and naturally precedes $T$ or $T'$.

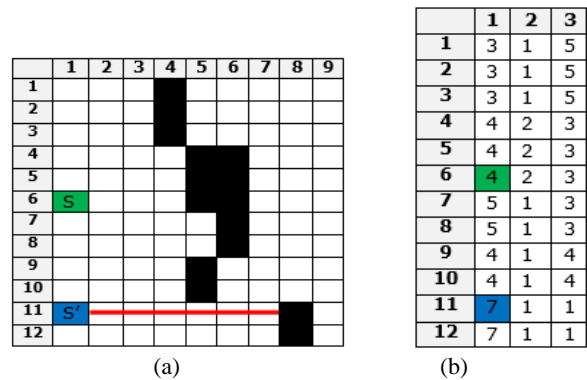

**Fig 9.** Visualization of a search space predicting the actual source node. Fig (a) represents the spatial coordinate of an image and Fig (b) represents its RLE format. The green tile is the initial start node and the blue tile is the actual start node. A red line represents the distance from source node to the hub.

**4.4. Handling Obstacles**

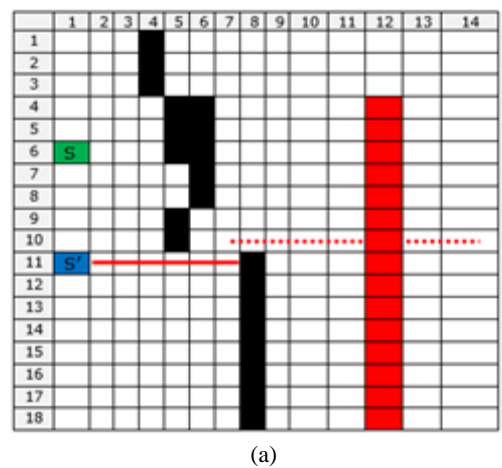

(a)

International Journal of Intelligent Systems and Applications in Engineering    IJISAE, 2018, 6(4), 251–261 | **256**

|   | 1 | 2 | 3 | 4 | 5 |
|---|---|---|---|---|---|
| 1 | 3 | 1 | 10 | 0 | 0 |
| 2 | 3 | 1 | 10 | 0 | 0 |
| 3 | 3 | 1 | 10 | 0 | 0 |
| 4 | 4 | 2 | 5 | 1 | 2 |
| 5 | 4 | 2 | 5 | 1 | 2 |
| 6 | 4 | 2 | 5 | 1 | 2 |
| 7 | 5 | 1 | 5 | 1 | 2 |
| 8 | 5 | 1 | 5 | 1 | 2 |
| 9 | 4 | 1 | 6 | 1 | 2 |
| 10 | 4 | 1 | 6 | 1 | 2 |
| 11 | 7 | 1 | 3 | 1 | 2 |
| 12 | 7 | 1 | 3 | 1 | 2 |
| 13 | 7 | 1 | 3 | 1 | 2 |
| 14 | 7 | 1 | 3 | 1 | 2 |
| 15 | 7 | 1 | 3 | 1 | 2 |
| 16 | 7 | 1 | 3 | 1 | 2 |
| 17 | 7 | 1 | 3 | 1 | 2 |
| 18 | 7 | 1 | 3 | 1 | 2 |

(b)

**Fig. 10.** Visualization of a search space tunneling an obstacle. Fig (a) represents the spatial coordinate of an image and Fig (b) represents its RLE format. The green tile is the initial start node and the blue tile is the actual start node. Blue tiles represent the intermediate hub and red tiles represents obstacle. A red line represents the distance from source node to the hub.

This section provides the working principle of handling obstacles. We encountered two types of obstacles while tunneling path as mentioned in corollary e. Generally, an obstacle occurs due to unconstrained writing style especially while scribing on an un-ruled paper. One of the possibilities is when two adjacent lines are touching one another at some positions. The other possibility would be when the ascenders or desceders of the characters extended beyond the search limit or range of the given threshold ($t$). For illustration purpose, we have considered an example which is shown in Figure 10(a) and Figure 10(b) representing spatial domain and its compressed version of a document image. We have chosen the threshold ($t = 6$) for demonstration. In this example, an initial source node is assumed, and it holds a position, $G(6,1)$. Previous section detailed about choosing appropriate starting node and applying that concept results in a new position, say $G(11,1)$ as a new Start Node. The new source node $G(11,1)$ has a maximum weight or distance (say 7 units) and falls within the range of $t$, which is closer to the initial source node $G(6,1)$. Further, the first successor hub is chosen as $G(10,3)$ which carries a weight of 6 units and covers the distance of 11 units from the left margin. It is chosen based on fact that the distance, say 11 units, is greater than the distance (weight) of source node that covers the distance of 7 units. The other facts include the search space (calculated distance) between the range $G(11 - t, *)$ and $G(11 + t, *)$ is either lesser or equal to the current position and this node aligns closer to its predecessor node along $y$ axis. Further, next hopping hub is complicated because we encounter an obstacle along the $y$ coordinate. Further, the chosen threshold value ($t$) is restricted within the search space. In this situation, we can crossover the hurdle by choosing the next hub as $G(10,4)$ which is identified as a successor hub.

### 4.5. Finding Correspondence between Source and Target

Identification of correspondence between the terminal points residing at opposite margins is one of the challenging aspects as mentioned in the study [13]. In this article, actual source node is chosen based on the weight (distance) which is distributed across its neighbors. The destination is based on the search space of the proposed model. The agent may start at a new source position and would reach the probable destination, and this articulates the correspondence between these two nodes present at extreme ends of the document. This is applicable for all the source nodes and target nodes. Figure 11 shows the correspondence between the source and terminal nodes and the text-line segmentation. In section 3 we have described the correspondence of the nodes notably having one-to-one relation and strictly no onto relationship and additionally no crossovers are allowed between the terminal points. This could be related to the bi-partite graph theoretical approach and thus maintaining the order or position.

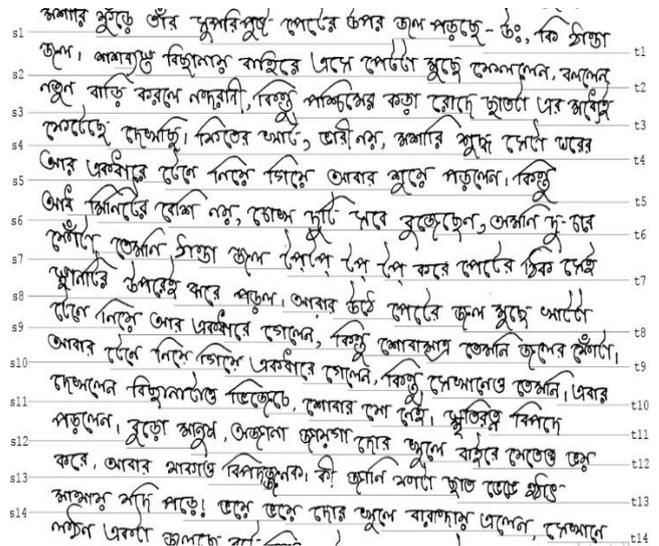

**Fig. 11.** Line Terminals S and T at both ends of a document in uncompressed documents (Reference: A portion of ICDAR13 test image - 309.tif)

The problem of over separation mentioned in the study [13] highlights the aspect of spotting two or more separator points within two adjacent text-lines. This occurs when a text line is identified as a non-text (white space) region. One of the reasons for over separation is because of concavity of the character. The other affecting factor could be multiple disjoint fractions or components which compose a character. Figure 12 depicts the problem of over separation where source nodes $s1$ and $s2$ start between two adjacent text-lines.

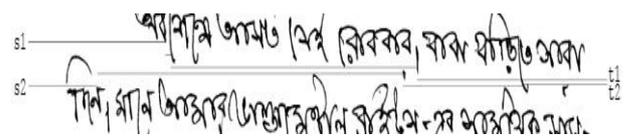

**Fig. 12.** Line Terminals S and T at both ends of a document in uncompressed documents (Reference: A portion of ICDAR13 test image - 309.tif)

Taking an average distance between the separator points may not produce the expected result as experimented in the research [13]. Our proposed method tackles this problem by understanding correlation between the adjacent paths. In this example, we have two source points such as s1 and s2 and the paths traced by the agent reaches the terminal points $t1$ and $t2$ respectively. It is evident that the terminal points $t1$ and $t2$ are very much closer to one another. Even both the paths traced are aligned together along $y$ coordinates most of the time. To resolve this issue, we need to ignore the path(s) which relate to more hubs (nodes). In other words, it is to retain the path which holds less intermediate hubs. It is also necessary to retain a path which is relatively parallel to both of its predecessor and successor paths.

Sometimes, this analogy may fail when a text line occupies less space when compared to its predecessors as shown in figure 13.

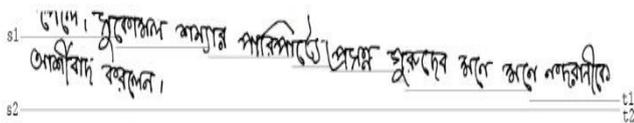

**Fig. 13.** Line Terminals S and T at both ends of a document in uncompressed documents (Reference: A portion of ICDAR13 test image - 309.tif)

This occurs mostly with a last text-line of a paragraph having less number of words. However, this situation is addressed by understanding the average distance between the two adjacent paths as mentioned above. It is computed by

$$r = \frac{\sum(u-v)}{n}$$

where $u = y$ coordinate postions in $G$
for the reference path with regular interval $t'$

where $v = y$ coordinate postions in $G$ for the successor path
with regular interval $t'$

$n =$ number of points

We illustrate the distance between the adjacent paths with an example. Figure 14 shows the paths traced along the two text-line gaps.

Table 4 shows the coordinate positions of paths traced along $y$ axis. We assume $t'$ as 500 for illustration purpose. For every interval $t'$, we take the $y$ coordinate positions for both paths. Table 5 shows $y$ coordinate positions for the paths with an interval gap of $t'$. The distance calculated for the paths (path 1 and path 2) is given under.

Given $n = 5$ and $t' = 500$,

Then $r = 63$

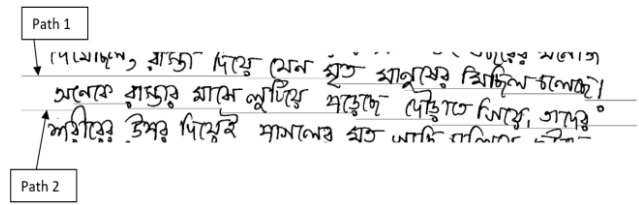

**Fig. 14.** Two paths traced along the text-lines

**Table 4.** RLE coordinate positions for two adjacent paths.

| Hubs | Path 1 | | Path 2 | |
|---|---|---|---|---|
| | Y-Coordinates | Distance covered from left margin | Y-Coordinates | Distance covered from left margin |
| Source | 1977 | 767 | 2079 | 186 |
| Hub 1 | 1984 | 1058 | 2059 | 367 |
| Hub 2 | 2000 | 1295 | 2073 | 682 |
| Hub 3 | 2020 | 1325 | 2087 | 1132 |
| Hub 4 | 2014 | 1626 | 2105 | 1389 |
| Hub 5 | 2029 | 1901 | 2123 | 1665 |
| Hub 6 | 2049 | 2085 | 2132 | 1870 |
| Hub 7 | - | - | 2124 | 2085 |

**Table 5.** RLE coordinate positions for two adjacent paths.

| Hubs | $t'$ | Y-Coordinates (Path 1), $u$ | Y-Coordinates (Path 2), $v$ |
|---|---|---|---|
| Source | 1-500 | 1977 | 2059 |
| Hub 1 | 501-1000 | 1977 | 2073 |
| Hub 2 | 1001-1500 | 2020 | 2105 |
| Hub 3 | 1500-2000 | 2029 | 2132 |
| Hub 4 | 2001-2500 | 2049 | 2124 |

## 5. Experimental Evaluation

### 5.1. Datasets

Our proposed method is evaluated on various benchmark handwritten datasets such as ICDAR'13 [31] and others [17] comprising of Kannada, Oriya, Persian and Bangla documents. For experimental purpose, the compression standards for these datasets are preserved as discussed in one of the studies [16].

### 5.2. Results

Our proposed method is evaluated based on two factors that we came across in a study [31] - (i) Detection Rate ($DR$) and (ii) Recognition Accuracy ($RA$). $DR$ and $RA$ are defined as follows:

$$DR = \frac{o2o}{N},$$

$$RA = \frac{o2o}{M},$$

where $o2o$ represent one to one matching and

$N$ be the count (separator points) of ground $-$ truth elements,

$M$ be the count of result elements

The DR in spotting the separator points at both the left side and the right side of the document is provided in literature [13] and also been tabulated (Table 6).

**Table 6.** Detection Rate

| Datasets (Handwritten) | Total Lines (N) | Detected | | | |
|---|---|---|---|---|---|
| | | o2o | | Rate (%) | |
| | | Left | Right | Left | Right |
| ICDAR13 [31] | 2649 | 2578 | 2502 | 97.31 | 94.45 |
| Kannada [20] | 4298 | 4173 | 4082 | 97.09 | 94.97 |
| Oriya [20] | 3108 | 3012 | 2911 | 96.91 | 93.66 |
| Bangla [20] | 4850 | 4650 | 4598 | 95.87 | 94.80 |
| Persia [20] | 1787 | 1690 | 1723 | 94.57 | 96.41 |

Table 7 shows the result of the proposed model applied on the benchmark datasets. Our method focuses only on a terminal point that is spotted on the left side of the document, so the RA for segmenting the text line, entirely depends upon the separator points spotted along left margin of the document. Figure 15 shows the comparative performance analyses of both DR and RA for various benchmark datasets. It is evident that the algorithm works better in the case of ICDAR datasets. We also witness that the RA for the dataset of Persian script is lower when compared to other datasets. This is because of the reason that the Persian writing style starts from right end of the document and precedes towards left direction unlike most of the other scripts. The other reasons include more concavity and disjoints in the composition of the character.

**Table 7.** Accuracy Rate

| Datasets | Total Lines (N) | o2o (Left) | Recognition Rate in % (Segmentation) |
|---|---|---|---|
| ICDAR13 | 2649 | 2578 | 89.2 |
| Kannada | 4298 | 4173 | 87.21 |
| Oriya | 3108 | 3012 | 85.00 |
| Bangla | 4850 | 4650 | 84.91 |
| Persian | 1787 | 1690 | 82.08 |

The algorithmic modeling deals with choosing a common threshold value ($t$) for every search space $SS$ to handle the obstacle $O$ is based on experimentation conducted on the datasets. Figure 16 shows the RA for different threshold values with respect to various datasets. The threshold value $t$ with 20 units provides better accuracy rate with respect to ICDAR13, Kannada, Oriya and Bangla. Whereas the threshold value $t$ possesses a unit value of 28 which elevates the accuracy rate in case of Persian script. In this research work, we have chosen the threshold value $t$ as 20 units which is common to all the datasets.

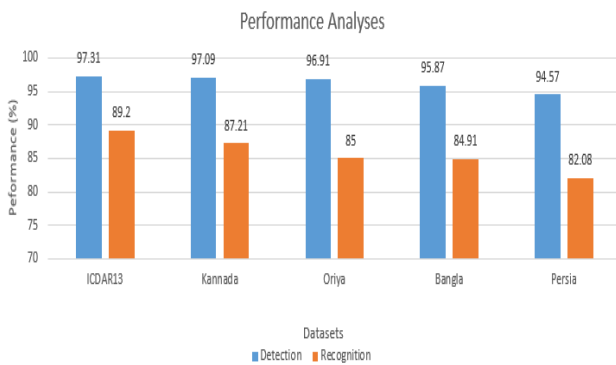

**Fig. 15.** Performance evaluation with respect to different datasets

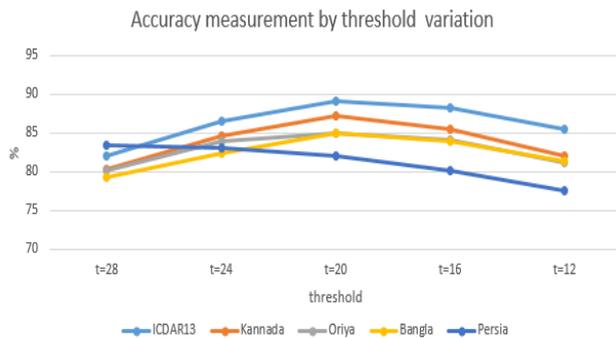

**Fig. 16.** Selection of threshold based for efficacy of the proposed model.

Real-time performance is measured employing the algorithm on various benchmark datasets to understand the working principle of the model. Figure 17 shows the comparative study of both (i) compressed and (i) uncompressed versions of dataset. Here, we witness that the processing time of compressed version of document takes lesser time unit (milliseconds) when compared to uncompressed version of documents. We also observe that there is an invariable exponential increase in the amount of time for performing text-line segmentation in case of uncompressed documents with increase in the number of images.

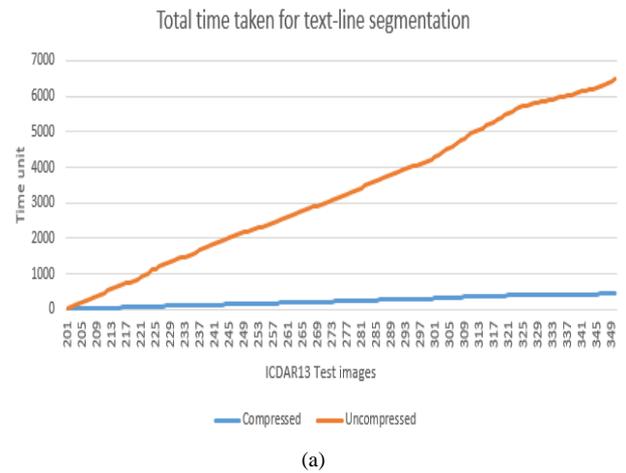

(a)

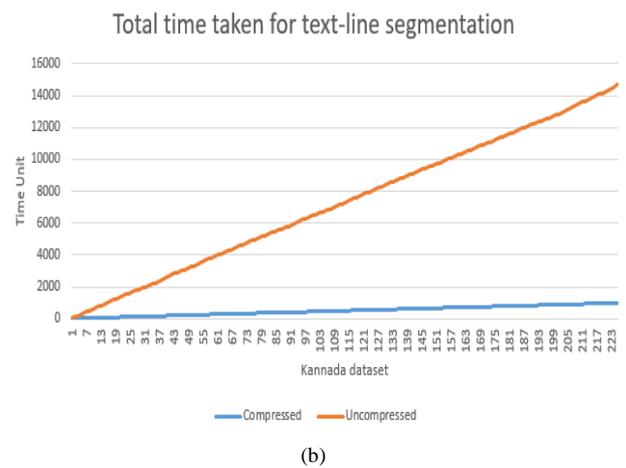

(b)

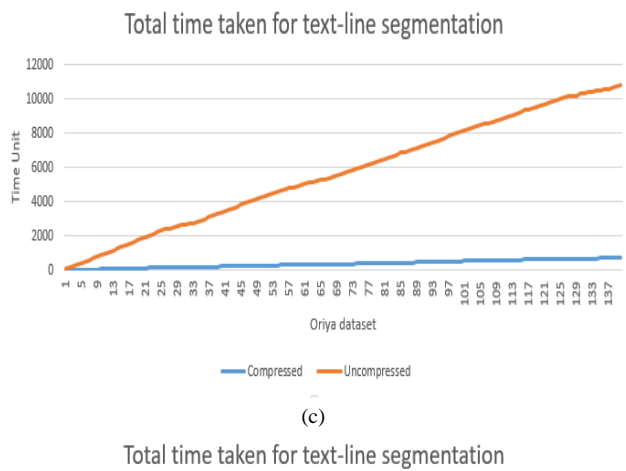

(c)

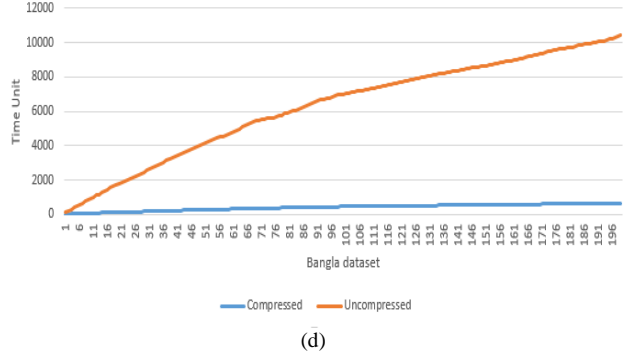

(d)

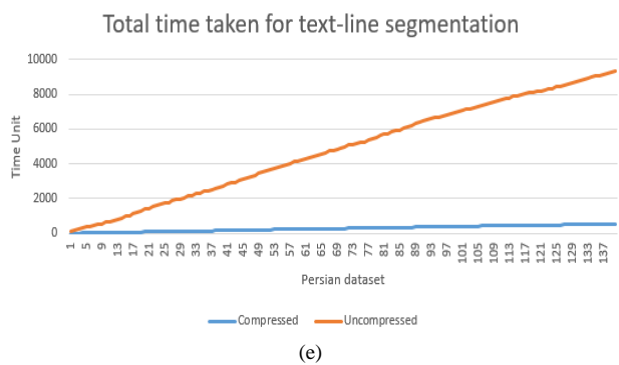

(e)

**Fig. 17.** Measurement of comparative real-time performance of both compressed and uncompressed versions of the images. Time Unit is given in milliseconds along y axis and x axis represents the sequence of images of each dataset.

Finding correspondence between $S$ and $T$ is a challenging task. However, we have attempted to find the correspondence based on the correlation between the adjacent paths as illustrated in Section 4.5. The first $P$ of the document starts from the very first terminal point identified in the first column of $G$. $P$ is presumed as a reference line. Now, the correlation is calculated between a new path and the reference line with an interval of $t'$. The same is repeated for the following paths with reference to source nodes. Figure 18 shows the RA because of the model.

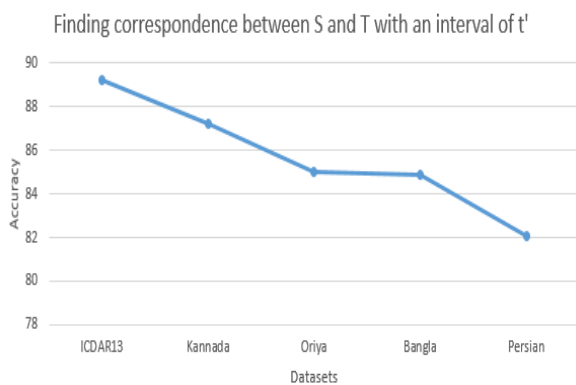

**Fig. 18.** RA after identifying the correspondence between the adjacent text-lines with an interval of t'.

## 6. Conclusion

In this paper, we have proposed an algorithm that segments the text-line of a document image by operating directly in its compressed representation. We make use of current state-of-the-art technology that identifies the terminal positions at every text-lines in the compressed representation. We have shown the working principle of the algorithm with an illustration. An agent (or robot) has been involved to tunnel the path between the terminal points spotted at the opposite extreme ends of the document representation. We have discussed the search space for tunneling the path when it encounters obstacles because of oscillation, tilt, touching and curvy text-lines. We have advocated the reasons for using the threshold values with respect to search space by experimental results. Comparative study of processing uncompressed and compressed versions for text-line segmentation is also been detailed. Further, a significant reduction in time complexity when compared to processing uncompressed image for segmentation is showcased. Some of the interesting observations are addressed to overcome the limitations [13] such as finding the correspondence between the terminal points. Interestingly, the tunnel or path traced that results in segmenting two adjacent text lines are entirely based on the terminal points. The source terminal points are predicted based on the search space, SS, whereas the corresponding target points are identified by the agent.

Though working directly on an uncompressed version of the document is very challenging, we have designed a model that operates directly on compressed representation of the document images for text-line segmentation. We could achieve the recognition accuracy of 89.2% using the benchmark dataset (ICDAR13). The confidence level in achieving 89.2% with respect to this dataset is 100%.

The proposed model has some of the limitations such as working with invariable skew or tilt levels. The accuracy rate depends upon the detection rate observed from the extensive literature and a common threshold value $t$ for entire datasets. Other limitation includes calculating the correlation between the paths based on reference line. One of the future avenues include working on bi-directional search where we employ two agents starting simultaneously from opposite terminal points and meets at a junction. Employing multi-agents leads to parallel processing which would improve the performance of the system. We have followed unguided medium to tunnel the path by adapting various algorithmic strategies including greedy and dynamic programming with AI techniques. We could possibly avoid wrong segmentation by using guided medium [4] along with the strategy which would be one of the future direction of this research work.